\documentclass[runningheads]{llncs}
\usepackage{graphicx}
\usepackage{booktabs}
\usepackage{tikz}
\usepackage{multirow}
\usepackage{multicol}
\begin{document}
\title{\textsc{LonXplain}: Lonesomeness as a Consequence of Mental Disturbance in Reddit Posts}
\author{Muskan Garg\inst{1,4},
Chandni Saxena\inst{2},
Debabrata Samanta\inst{3},
Bonnie J. Dorr\inst{4}
}
\authorrunning{Garg et al.}
\institute{Mayo Clinic, Rochester, MN, USA\\
\email{muskanphd@gmail.com}\\
\and
The Chinese University of Hong Kong, Hong Kong, SAR\\
\email{chandnisaxena@cse.cuhk.edu.hk}\\
\and
Rochester Institute of Technology, Pristina, Kosovo, Europe\\
\email{debabrata.samanta369@gmail.com}
\and
University of Florida, Gainesville, FL, \\
\email{bonniejdorr@ufl.edu}
}
\maketitle    
\begin{abstract}
Social media is a potential source of information that infers latent mental states through Natural Language Processing (NLP). While narrating real-life experiences, social media users convey their feeling of loneliness or isolated lifestyle, impacting their mental well-being. Existing literature on psychological theories points
to \textit{loneliness} as the major consequence of \textit{interpersonal risk factors}, propounding the need to investigate loneliness as a major aspect of mental disturbance. We formulate \textit{lonesomeness detection} in social media posts as an \textbf{explainable} binary classification problem, discovering the users at-risk, suggesting the need of resilience for early control. To the best of our knowledge, there is no existing explainable dataset, i.e., one with human-readable, annotated text spans, to facilitate further research and development in loneliness detection causing mental disturbance~\cite{garg2023mental}. In this work, three experts: a senior clinical psychologist, a rehabilitation counselor, and a social NLP researcher define annotation schemes and perplexity guidelines to mark the presence or absence of lonesomeness, along with the marking of text-spans in original posts as \textit{explanation}, in $3,521$ Reddit posts. We expect the public release of our dataset, \textsc{LonXplain}, and traditional classifiers as baselines via GitHub.\footnote{\url{https://github.com/drmuskangarg/lonesomeness\_dataset}}.
\end{abstract}
\section{Introduction}
According to the \textit{World Health Organization},\footnote{\url{https://www.who.int/teams/social-determinants-of-health/demographic-change-and-healthy-ageing/social-isolation-and-loneliness}} one in three older people feel lonely. Loneliness 
%have 
has a serious impact on older people’s physical and mental health, quality of life, and their longevity. According to the \textit{Loneliness and the Workplace: 2020 U.S. Report}, three in five Americans
(61\%) report the feeling of loneliness, compared to more than half (54\%) in 2018~\cite{nemecek2020loneliness}. Older adults are at high risk for morbidity and mortality due to 
%experience 
prolonged isolation, especially during the 
%times of 
COVID-19~\cite{donovan2020social}
era. To this end, researchers demonstrate 
%the 
loneliness as a major concern for increased risk of depression, anxiety, and stress thereby affecting cognitive functioning, sleep quality, and overall well-being~\cite{cacioppo2006loneliness}. We define \textit{lonesomeness} as an unpleasant emotional response to perceived isolation through mind and character, especially in terms of concerns for social lifestyle. The anonymous nature of Reddit social media platform provides an opportunity for its users to express their thoughts, concerns, and experiences with ease. %To this end, we 
We leverage Reddit to formulate 
%the 
a new annotation scheme and perplexity guidelines for constructing an explainable annotated dataset for Lonesomeness. We start with an example of how this information is reported in social media texts.
\begin{quote}
    \textit{Example}: Within the last month, I have lost my best friend and my grandmother, to whom I was very close with (\textcolor{red}{both passed away}). This time last year, I was involved in an incident where I was \textcolor{blue}{assaulted} (we were not dating at the time), and it is bringing up some \textcolor{blue}{bad memories} for me. I am continuously mentioning about moving cities on my own. I am \textcolor{red}{all by myself} now and have \textcolor{red}{no one to speak to}!
\end{quote}

In %an example given above, 
this example, a person is upset about losing all loved ones and %handling 
coping with the memories of manipulative situations 
%of 
in their life. 
%We find the r
Red colored words depict
%ing the 
explanations for lonesomeness and blue colored words represent the triggering circumstances.

\begin{figure}[ht]
    \centering
    \includegraphics[width=0.78\textwidth]{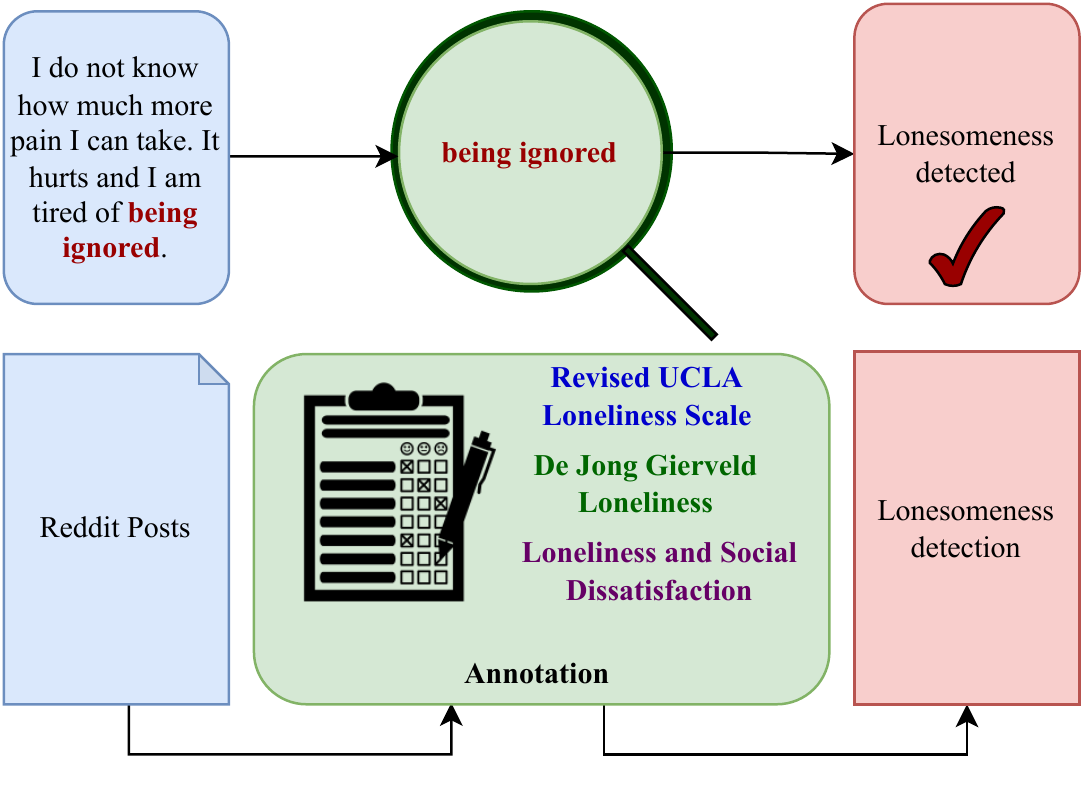}
    \caption{Overview of classifying lonesomeness in Reddit Posts.}
    \label{fig:overview}
    % \vspace{-0.35cm}
\end{figure}

% \subsection{Growing Popularity of Mental Health Forum on Reddit}
Sociologists
%, 
Weiss \textit{et. al.,} in 1975 \cite{weiss1975loneliness} introduce
%d 
a \textit{theory of loneliness} suggesting the need for six social needs to prevent loneliness in stressful situations: (i) attachment, (ii) social integration, (iii) nurturing, (iv) reassurance of worth, (v) sense of reliable alliance and (vi) guidance. Furthermore, Baumeister \textit{et. al.,} \cite{baumeister2017need} explains various indices of social isolation associated with suicide as living alone, and low social support across the lifespan. 
%On 
In recent investigations 
%with 
of
attachment style, Nottage \textit{et. al.,} \cite{nottage2022loneliness} argue
%s 
that loneliness mediates a positive association between attachment style and depressive symptoms. Loneliness results in disrupted work-life balance, emotional exhaustion, insomnia and depression \cite{becker2022surviving}. With this background, the annotation
%s 
guidelines are developed through the collaborative efforts of three experts (a clinical psychologist, a rehabilitation counselor and a social NLP researcher) for early detection of lonesomeness, which if left untreated, may cause chronic disease such as self harm or suicide risk.

% \muskan{VMHAs}
% \muskan{Need of KiL while moving from low level analysis to high level analysis}

% \muskan{Growing popularity of Mental health forum on Reddit: \url{https://arxiv.org/pdf/2206.00856.pdf} Figure 1}

%We examine the mental disturbance 
We examine potential indicators of mental disturbance
in Reddit posts, aiming to discover users at risk through 
explainable lonesomeness 
annotations, 
as shown in Figure~\ref{fig:overview}. %In this work, we 
We first introduce the Reddit dataset for lonesomeness detection in social media posts reflecting mental disturbance. 
%As a starting point for our study, this work 
Our data annotation scheme is designed to 
facilitate 
%s 
the discovery of 
%underlying 
users with 
%prospective 
(potential) underlying tendencies toward
%suicidal tendencies and 
self harm, including suicide, through loneliness detection. 
We construct this scheme
%The data annotation scheme is 
%them 
%constructed 
using three clinical questionnaires: (i) UCLA Loneliness Scale~\cite{russell1980revised}, (ii) De Jong Gierveld Loneliness Scale \cite{penning2014measuring}, and (iii) Loneliness and Social Dissatisfaction scale~\cite{asher1984children}. 

% \muskan{Add motivation: The need of contextual information via interpersonal risk of lonesomeness for Motivational Interviewing through VMHAs.}

% \muskan{NLP for mental health and therapy (Section 2) \url{https://arxiv.org/pdf/2106.01702.pdf}} 

\begin{table*}[]
    \centering
    \caption{Historical evolution of language resources for determining lonesomeness in texts. There is no existing publicly available or explainable dataset for identifying loneliness as an interpersonal risk factor for mental disturbance. }
    \begin{tabular}{p{4cm}p{1.5cm}|p{1.5cm}p{1.2cm}p{1cm}}
        \midrule \midrule
      
        \textbf{Dataset} &  \textbf{Media} & \textbf{Size} & \textbf{Xplain} & \textbf{Avail.}\\
        \midrule \midrule
        Kivran \textit{et. al.,} 2014~\cite{kivran2014understanding} & Twitter & 4454  & No  & No \\ 
        Badal \textit{et. al.,} 2021~\cite{badal2021words} & Interviews & 97 adults  & No  & No \\
        
        Mohney \textit{et. al.,} 2019~\cite{mahoney2019feeling} & Twitter & 22477 & No & No \\
        \midrule \midrule
        Ours & Reddit & 3522 & Yes & Yes\\
         
        \midrule \midrule
    \end{tabular}
    
    \label{tab:1}
    \vspace{-0.35cm}
\end{table*}

The quantitative literature on loneliness and mental health has limited, openly available language resources
%and its availability 
due to the sensitive nature of the data. Examples are
%as
shown in Table \ref{tab:1}. 
%Thus, we
We aim to fill this gap by 
%introduce 
introducing a 
new 
dataset for lonesomeness classification with human-generated explanations and to make it publicly available on Github. Our \textbf{\textit{contributions}} %can be 
are 
summarized as follows:

\begin{itemize}
    \item We %frame 
    define an experts-driven annotation scheme for Lonesomeness detection.

     \item We deploy the annotation scheme to construct and release \textsc{LonXplain}, a new dataset 
     %of 
     containing 
     $3521$ instances for early detection of textit{lonesomeness}
     that thwarts belongingness and 
     %prospective 
     potentially leads to self harm.
%XXX on social media by leveragingXX.

    \item We deploy 
    %the 
    existing classifiers and investigate explainability through Local Interpretable Model-Agnostic Explanations (LIME), suggesting 
    %the room for improvement towards 
    an initial step toward
    more responsible AI models.

%     % \item We 

% \item Experiments with traditional state-of-the-art methods and recent transformer-based models suggest room for improvement towards context-sensitive AI models. 

\end{itemize}

\section{Corpus Construction}
We collect Reddit posts through The Python Reddit API Wrapper (PRAW) API, from 02 December 2021 to 06 January 2022
%where we maintain the consistency of 
maintaining a consistent flow of
100 posts per day. The subreddits 
%that we use for constructing 
extracted for
this dataset are 
%the 
those
most widely used
in  % ??
the discussion forum for depression (\texttt{r/depression}) and suicide risk (\texttt{r/SuicideWatch}). We 
manually 
filter out 
%the 
irrelevant posts with empty strings and/or 
%the 
posts containing only URLs.
%, manually 
%and 
%recommend the same 
%serve up for further annotation. 

We further clean and preprocess the dataset and filter out the data samples 
(posts) longer than 300 words,
%BJD: I wasn't sure if you meant characters or words?  Also, I inserted "(posts)" above, to clarify what you mean by sample.
%with length $>$ 300, 
to simplify the complexity of a given task. The 
length of a single sample in the
original dataset varies from 1 to more than 4000 words,
%in a single sample,
highlighting the need 
%of upper limits in length, 
for bounding the length, 
%in order to develop AI models for comparatively consistent data points. 
thus inducing comparatively consistent data points
for developing AI models.
We 
%frame 
define
the experts-driven annotation scheme, train and employ three student annotators for data annotation and 
%carry out 
compute
inter-annotator agreement to ensure the coherence and reliability of the annotations. We emphasize FAIR principles~\cite{wilkinson2016fair} while constructing and releasing \textsc{LonXplain}.

\subsection{Annotation Scheme}
%URGENT: Who is "Dunn"?  It pops up in 2.1 without a citation.  I searched the whole doc and couldn't find it, even though there are Dunn entries in the .bib file. Did something get deleted?
Dunn introduces
%d 
six dimensions of wellness (spiritual, social, intellectual, vocations, emotional, and physical), affecting users' mental well-being. 
%%BJD: Starting to have trouble right about here in the text:
%
%One of the key consequence of these mental disturbance is evolved as loneliness
%
%(1) You say "these mental disturbance"--but those are dimensions of WELLNESS, not DISTURBANCE?
%(2) Are you saying a key consequence of WELLNESS is evolution of LONELINESS?  I think you need to say something abut how these dimensions can either be positive or negative, e.g., loneliness perhaps derives from negative values associated with the spiritual, social, and emotional dimensions?
%(3) Grammar: please watch plural/singular, a singular noun (e.g., "disturbance") takes a singular determiner ("this" not "these"). 
%(4) Doing my best to rephrase here, but please check it:
A key consequence of mental disturbance is a tendency toward the negative end of the scale for these dimensions, e.g., loneliness derives from negative values associated with the spiritual, social, and emotional dimensions. 
%further intensifies it to 
Further intensification may lead to
thwarted belongingness~\cite{ghosh2022no},
%and 
suicidal tendencies, and self harm. 

%BJD: Below it looks like you are trying to compensate and include balance between NLP and clinical psych, but you are doing it by introducing terms like trade-off and negotiating.  I did my best to rephrase this to make it more understandable and positive.
%The experts opinion on using Natural Language Processing (NLP) challenges and clinical questionnaires on loneliness detection are two concrete baselines for negotiating trade-off between this interdisciplinary approach of NLP and clinical psychology.
%BJD: Rephrased here:
Bringing together the two disiplines of Natural Language Processing (NLP) and clinical psychology, we adopt an annotation approach that leverages both the application of NLP to Reddit posts and clinicial questionnaires on loneliness detection, as two concrete baselines.
% \muskan{Why these questions are important? Rising from the Motivational Interviewing}
The annotations are based on two research questions: 
%(i) "RQ1: 
(i) ``RQ1:
\textit{Does the text contain indicators of lonesomeness which alarms suicidal risk or self harm in a person?},'' and 
%(ii) "RQ2:
(ii) ``RQ2: 
\textit{What should be the extent to which annotators are supposed to read in-between-the-lines for marking the text-spans indicating the presence or absence of lonesomeness}.''

Our experts 
%use 
access
three clinical questionnaires used by
%the 
mental health practitioner,
%in practice, to
%frame 
to define lonesomeness annotation guidelines.
%the 
%lonesomeness. 
The UCLA Loneliness Scale~\cite{russell1980revised}
%is used for measuring 
that measures loneliness % through 20 items 
%which 
was
%further revised to
adapted to distinguish among
%determine 
three dimensions of loneliness: \textit{social loneliness} (the absence of a social network), \textit{emotional loneliness} (the absence of a close and intimate relationship), and \textit{existential loneliness} (the feeling of being disconnected from the larger world). The De Jong Gierveld Loneliness Scale \cite{penning2014measuring} is a 6-item self-report questionnaire over 5-point Likert scale (ranging from 0 (not at all) to 4 (completely)) that assesses two dimensions of loneliness: \textit{emotional loneliness} and \textit{social loneliness}. 

From Loneliness and Social Dissatisfaction scale~\cite{asher1984children} we use 10 out of 20 items, reflecting loneliness on a 5-point Likert scale, ranging from 1 (strongly disagree) to 5 (strongly agree). The experts annotate 40 data points using fine-grained guidelines seperately at 3 different places to avoid any influence. Furthermore, we find the possibility of dilemmas due to the psychology-driven subjective and complex nature of the task.

\subsection{Perplexity Guidelines}
We propose 
%the 
perplexity guidelines to simplify the task and facilitate future annotations. We observe following:
%major confusions:
% \muskan{Rephrase language below}
\begin{enumerate}
    \item \textbf{Lonesomeness in the Past}: %To check if
    A person
    %with lonely past 
    with a history of loneliness
    may still be 
    %alarming prospect of 
    at risk of self harm or suicide.
    %al risk. 
    For instance, '\textit{I was so upset being lonely before Christmas and today I am celebrating New Year with friends'}. We %frame
    define
    rules to 
    %handle 
    capture indicators of prior lonesomeness
    %in the past 
    %because a person attends celebration and 
    such as attending a celebration to fill the void
    associated with this negative emotion.
   % overcome the preceding mental disturbance which means filling void with external event. 
   With both negative and positive clauses in the example above, the NLP expert would deem this neutral, yet our 
   %neutral opinion by NLP expert about double negation, our 
   clinical psychologist 
   %argues 
   discerns
   the presence of lonesomeness, with both clauses contributing to its likelihood.  
   %in their perception which may again evolve after some time 
   This post is thus marked
   %with the presence of 
   as presenting lonesomeness, 
   an indicator that the author is potentially at risk.
   %as thus, user at-risk.
    \item\textbf{Ambiguity with \textit{Social Lonesomeness}}: %Relationships point to the importance of the ability to take a societal pulse on a regular basis, especially in these unprecedented times of pandemic-induced distancing and shut-downs. 
    %People mention major 
    Major societal events such as breakups, marriage, best friend related issues may be mentioned in different contexts, suggesting different perceptions.
    %in user perceptions. 
    We formulate two annotation rules:
    %mitigate this problem with two statements: 
    (i) Any feeling of void/ missing/ regrets/ or even mentioning such events with negative words %should be 
    is 
    marked as the presence of lonesomeness. Example:
    %such as 
    %\textit{'But 
    \textit{`But
    I just miss her SO. much. It's like she set the bar so high that all I can do is just stare at it.'}, 
    (ii) %Anything associated with 
    Mentions of fights/ quarrels/ general stories %should be 
    are 
    marked with absence of lonesomeness. Example:
    %such as 
    %\textit{'My
    \textit{`My
    husband and I just had a huge argument and he stormed out. I should be crying or stopping him or something. But I decided to take a handful of benzos instead.'}. 
    
\end{enumerate}

\subsection{Annotation Task}

% \muskan{Rephrase below}
We employ three postgraduate students, trained by experts on manual annotations.
%a given 
%The p
Professional training and guidelines are supported by perplexity guidelines. After three successive trial sessions to annotate 40 samples in each round, we ensure their \textbf{coherence} and understanding of the task for further annotations. 

Each data sample is annotated by three annotators in three different places to confirm the \textbf{authenticity} of the task. We restrict the annotations to 100 per day, to maintain the \textbf{quality} of the task and \textbf{consistency}. We further validate three annotated files using Fliess' Kappa inter-observer agreement study to ensure \textbf{reliability} of the dataset, where kappa is calculated 
at
%as 
$71.83\%$, and carry out agreement studies for lonesomeness detection. 
We obtain final annotations based on 
%the 
a \textit{majority voting mechanism} and experts' opinions, resulting in \textsc{LonXplain} dataset. 
Furthermore, the explanations are annotated by a group of 3 experts to ensure the nature of \textsc{LonXplain} as \textit{psychology-grounded} and \textit{NLP-driven}. We deploy FAIR principle~\cite{wilkinson2016fair} by releasing \textsc{LonXplain} data in a public repository of Github, making it \textbf{findable} and \textbf{accessible}. The comma separated format contains $<$text, label, explanations$>$ in English language, ensuring the \textbf{interoperability} and \textbf{re-usability}. We illustrate the samples of \textsc{LonXplain} in Table~\ref{tab:4}, with blue and red color indicating the presence of \textit{cause}~\cite{garg2022cams} and \textit{consequence}~\cite{ghosh2022persona}, respectively. This task of early consequence detection, may prevent chronic disease such as depression and self-harm tendencies in the near future.

\begin{table*}[]
    \centering
    \caption{An annotated dataset example illustrates
    %sample of dataset to examine 
    causes (blue colored text-span) and lonesomeness as a consequence (red colored text-span) of mental disturbance in Reddit posts. 
    %All the posts may not 
    Not all posts contain information about cause and/or consequences.}
    \begin{tabular}{p{9cm}|p{1.2cm}|p{1.4cm}}
        \midrule
      
        \textbf{Text}  & \textbf{Label} & \textbf{Exp.}\\
        \midrule
        Just a sense of impending doom, this year is going to be shit. I'm starting to think things never actually do get better. All of \textcolor{red}{my friends are out partying right now }$\leftarrow$ (Consequence) and I'm at home \textcolor{blue}{getting lectured by my family}$\leftarrow$ (Cause) on my negative attitude. Anyway, happy new year I guess. & Present & my friends are out partying\\
        \midrule
        All of us on here are probably \textcolor{red}{feeling alone and lonely}$\leftarrow$ (Consequence) and depressed and like \textcolor{blue}{everyone else out there is having an awesome time}$\leftarrow$ (Cause) except us, so why don't we have our own "party"? (In a way). Let's get to know each other! What is something really funny to you guys? It can be a joke/a meme/a video/a story of yours/whatever. Let's help each other feel less alone. & Present & feeling alone and lonely \\
        \midrule
        There is literally no point in life. We live and we die. And \textcolor{blue}{life has been hell to me}$\leftarrow$ (Cause) so far so why should I even bother finishing. I am almost at the point where I am about to say **** it and quit. & Absent & -\\

         \midrule
    \end{tabular}
    
    \label{tab:4}
\end{table*}

\section{Data Analyses}
%The c
Corpus construction is accompanied by fine-grained analyses for: (i) statistical information about the dataset and
%nature of the dataset through statistical information, and 
(ii) overlapping terms and syntactic similarity based on
%through 
word cloud and keyword extraction. In this section, we further discuss the linguistic challenges with supporting psychological theories for \textsc{LonXplain}.

\subsection{Statistical Information}
\textsc{LonXplain} contains $3,521$ data points among which 54.71\% are labeled as positive sample, depicting the presence of lonesomeness in a given text. We observe the statistics for number of words and sentences in both the \texttt{Text} and \texttt{Explanation} (see Table \ref{tab:2}). The average number of Text words
%are
is
4 and the maximum number of words 
%are 
reported as explainable in text spans is 19,
%in explainable text-spans, 
highlighting the need %of 
for
identifying focused words for classification.
\begin{table}[]
    \centering
    \caption{The statistics of \textsc{LonXplain} for determining the presence or absence of lonesomeness in a given Reddit post.}
    \begin{tabular}{c|ccc}
        \midrule \midrule
        \textbf{Column} &\textbf{Feature} & \multicolumn{2}{c}{\textbf{Lonesomeness}}\\
       
      &  & Absent & Present\\
        \midrule \midrule
        % \textbf{Statistics}\\
        Labels & Number of Posts & 1595 & 1927\\
        \midrule
      Text %& Minimum number of Words & 4 & 1\\
       & Average number of Words & $\approx$97 & $\approx$135\\
       & Maximum number of Words & 300 & 300 \\
       & Total number of Words & 153459 & 258992 \\
        % & \midrule & \midrule & \midrule\\
      % & Minimum number of Sentences & 1& 1 \\
       & Average number of Sentences & $\approx$7 & $\approx$9 \\
        & Maximum number of Sentences & 42 & 32 \\
        & Total number of Sentences & 9618 & 16302 \\
        \midrule
          Explanation %& Minimum number of Words & - & 1\\
        & Average number of Words & - & $\approx$4\\
        & Maximum number of Words & - & 19 \\
        & Total number of Words & - & 6647 \\
    \midrule \midrule
    \end{tabular}
    
    \label{tab:2}

\end{table}

\subsection{Overlapping Information}

\begin{figure*}[]
    \includegraphics[width=.31\textwidth]{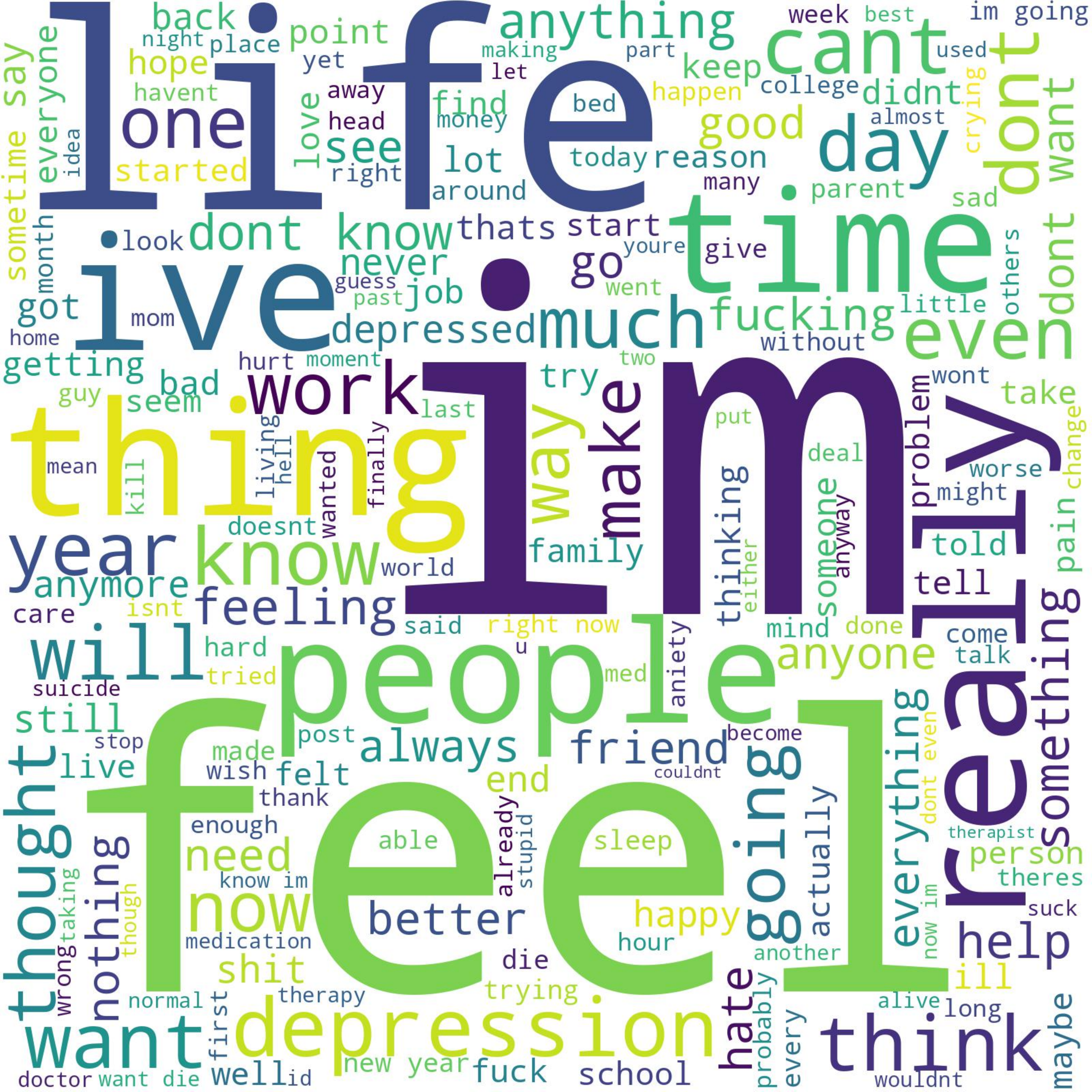}\hfill
    \includegraphics[width=.31\textwidth]{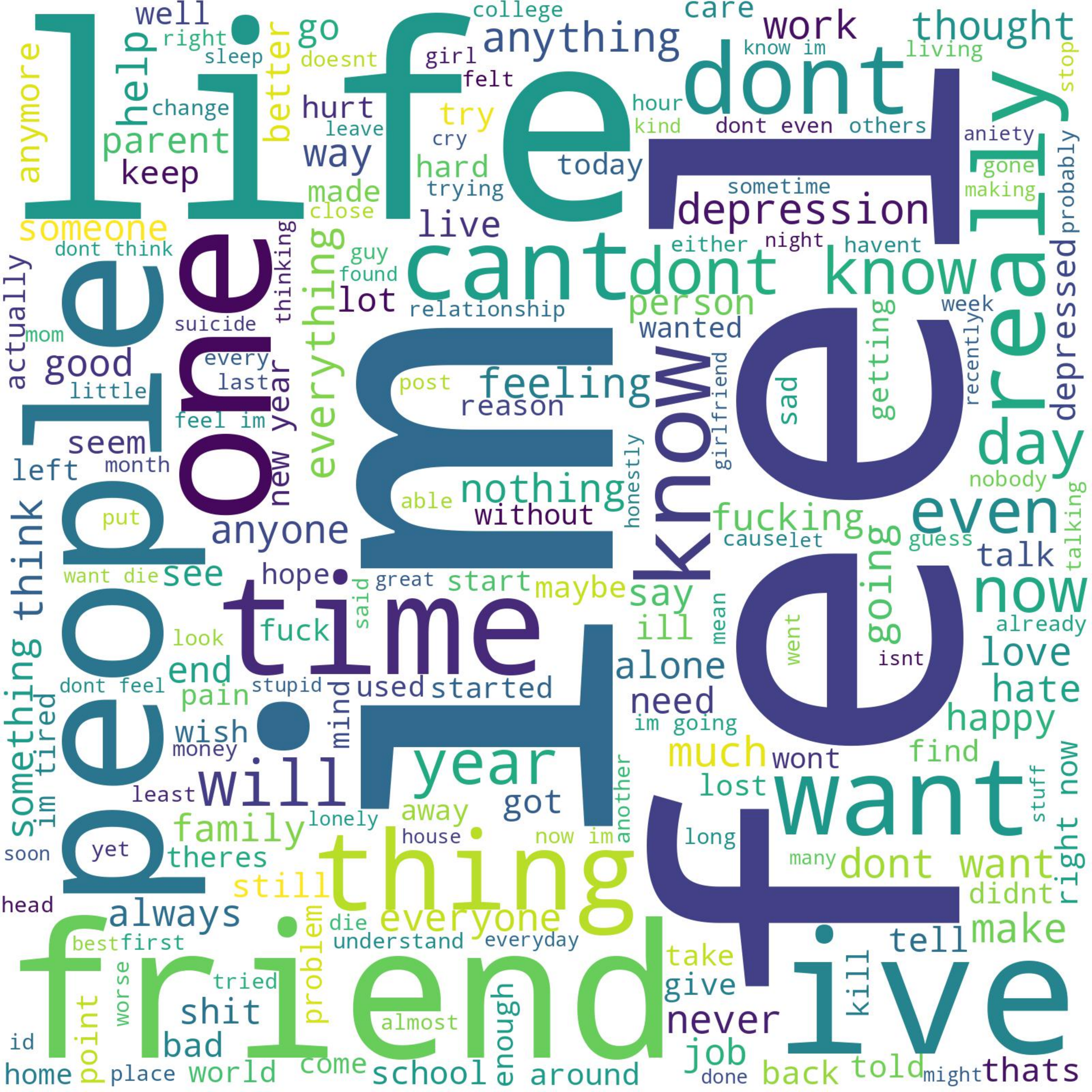}\hfill
    \includegraphics[width=.31\textwidth]{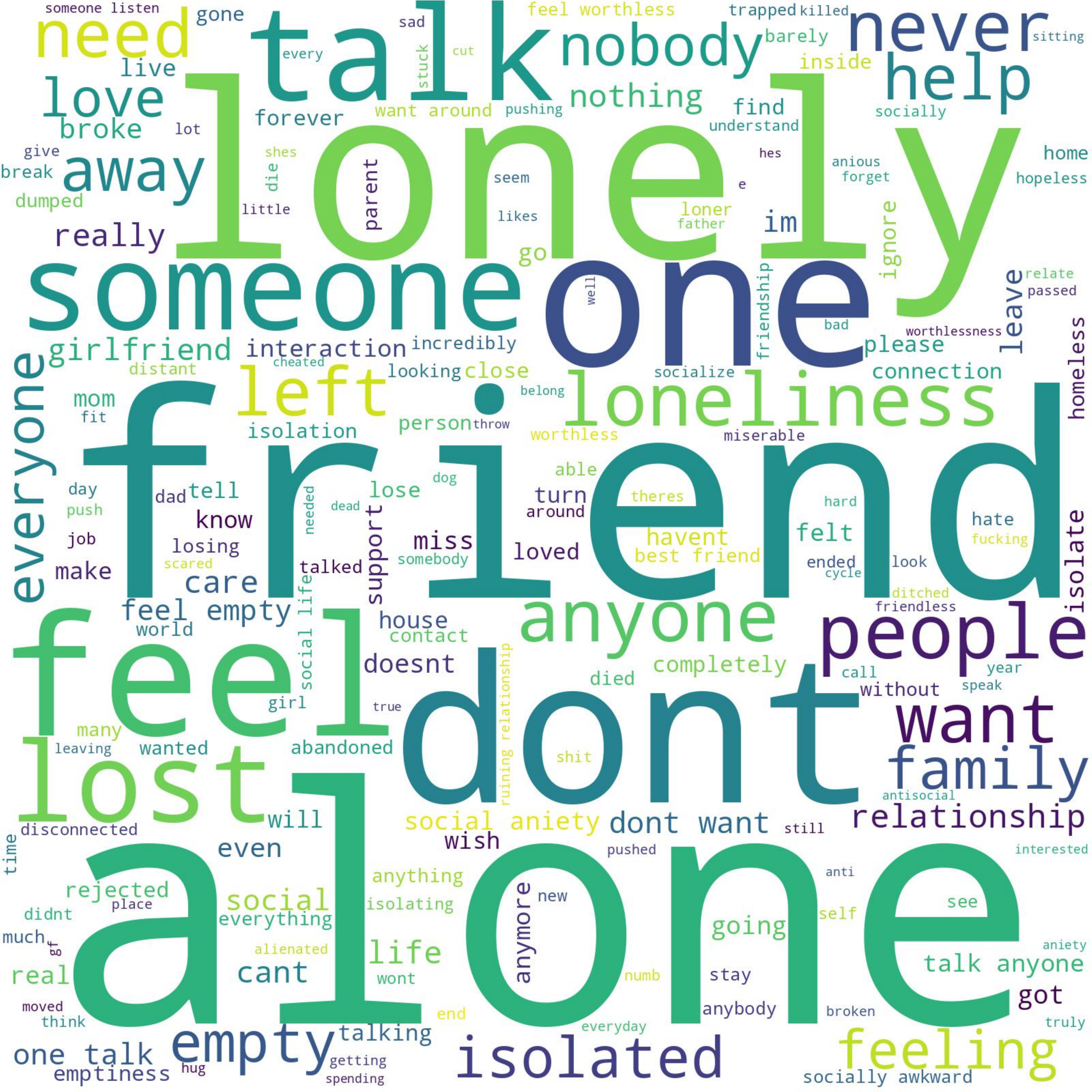}

    \caption{The wordcloud for label 0: absence of lonesomeness (left), label 1: presence of lonesomeness (center), and Word cloud evolved from explainable text-spans (right)}
    \label{fig:wordcloud}

\end{figure*}

We investigate %the 
words that represent class 0: absence of lonesomeness and class 1: presence of lonesomeness in a given text. Words such as \texttt{life, im, feel, and people} indicate 
%huge 
a very large syntactic overlap %among both the classes 
between these classes (see Fig.~\ref{fig:wordcloud}). However, a
%%BJD: I'm strengthening this:
seemingly neutral (or even positive) word like
\texttt{friend} indicates 
%the discussion 
a discussion about interpersonal relations, which 
%might be a reason of impacting mental disturbance. 
be an indicator of a negative mental state.
We further obtain word clouds for explainable text-spans indicating label 1 in \textsc{LonXplain}. 
%We found words like 
We find that words such as \texttt{lonely, alone, friend, someone, talk} 
%stipulates the context to identify lonesomeness in a given text. 
may be indicators of the presence of lonesomeness.

%We use keyword extraction mechanism: 
For keyword extraction, we use KeyBERT, a pre-trained model that finds the sub-phrases in a data %point 
sample
reflecting the semantics of the original text.  
%the 
BERT extracts \textit{document embeddings} 
%are extracted with the BERT model to get 
as a document-level representation and 
%Next, followed by the 
\textit{word embeddings} for N-gram words/phrases~\cite{grootendorst2020keybert}. Consider a top-20 %list of top 20 words 
word list for each of label 0 ($K0$) and label 1 ($K1$):
%respectively
\begin{quote}
    $K0:$ \textit{sleepiness, depresses, tiredness, relapses, fatigue, suffer, insomnia, suffers, stressed, sleepless, selfobsessed, stress, numbness, cry, diagnosis, sleepy, moodiness, depressants, stressors, fatigued}.\\
    
    $K1:$ \textit{dumped, hopelessness, introvert, breakup, hopeless, loneliness, heartbroken, dejected, introverted, psychopath, breakdown, graduation, breakdowns, overcome, depressant, solace, counseling, befriend, sociopaths, abandonment}. 
    
\end{quote}

Although 
%there are
some important terms 
%that 
are missed by KeyBERT, 
e.g.,
%such as 
\texttt{homelessness} and \texttt{isolated},
%to name a few, 
most of the terms in $K1$ %are indicators of 
indicate lonesomeness. Thus, KeyBERT plays a pivotal role in %marking keywords depicting the need of 
LonXplain, as an example of a context-aware AI classifier 
%s, thereby making them more 
that lends itself to explainable output, and (more generally) responsible AI.

\section{Experiments and Evaluation}
We formulate the problem of lonesomeness detection as a binary class classification problem, define the performance evaluation metrics, and discuss the existing classifiers. Following this, we 
%brief out 
present
the experimental setup and implement the existing classifiers for setting up baselines on \textsc{LonXplain}. The explainable AI method, LIME~\cite{nguyen2018comparing}, is used to find text-spans responsible for decision making, highlighting the scope of improvement.

\paragraph{Problem Formulation.}
We define the task of identifying Lonesomeness $L$ and its explanation $E$ in a given document $D$. The ground-truth contains a tuple $<D (text), L (bool), E (text)>$ for every data point in a comma separated format. A corpus of D documents where $D=\{d_1, d_2, ..., d_n\}$ for $n$ documents where n=3,522 in \textsc{LonXplain}. We develop a binary classification model for every document $D_i$ to classify it as $L_i$.

\subsection{Experimental Setup}
\paragraph{Evaluation Metrics.}
We evaluate the performance of our experiments in terms of precision, recall and f-score.
%Precision represents the proportion of correctly identified lonely Reddit posts among all posts identified as lonely by the classifier. Recall would represents the proportion of correctly identified user's lonely posts among all truly user's lonely posts in the dataset. F1-score would represent the balance between correctly identifying user's lonely posts while minimizing the number of false positive identifications. We first obtain true positive, true negative, false positive, and false negative, followed by comparing the predicted outputs to true labels in \textsc{LonXplain}. 
%Accuracy is the proportion of correctly classified posts (both true positives and true negatives) out of all posts in \textsc{LonXplain}. It 
Accuracy is a simple and intuitive measure of overall performance that is easy to interpret. However, accuracy alone may not be a good measure of performance for imbalanced datasets, where one class (e.g., user's lonely posts) is not equal to the other (e.g., user's non-lonely posts). In such cases, a model that always predicts the majority class can achieve high accuracy, but will not be useful for detecting the minority class. In this work, our task is to identify lonesomeness and not to identify a user's non-lonely posts. Thus, we consider Accuracy as an important metrics for evaluation.

\paragraph{Baselines.}
We compare results with
% [leftmargin=*]
% \begin{itemize} 
linear classifiers using word embedding Word2Vec~\cite{NIPS2013_5021}. We further use GloVe~\cite{pennington2014glove} to obtain word embeddings and deploy them on two Recurrent Neural Networks (RNN): Long Short Term Memory (LSTM) and Gated Recurrent Unit (GRU) for performance evaluation.
\begin{enumerate}
    \item \textbf{LSTM:} 
    %The
    Long Short-Term Memory networks (LSTM) take a sequence of data as an input and make predictions at individual time steps of the sequential data. 
    We apply 
    a 
    LSTM model for classifying texts 
    %with 
    that indicate lonesomeness versus those that do not.
    %from those data samples which does not.
    \item \textbf{GRU:} %A 
    Gated Recurrent Units (GRU) 
    %intends to 
    use connections between the sequence of nodes to resolve 
    %a 
    the \textit{vanishing gradient problem}. 
    Since the textual sequences in the \textsc{LonXplain} present a mixed context, GRU's non-sequential nature offers improvement over LSTM. 

\end{enumerate}
%%BJD: Please check that I am getting this right:
We additionally
%Furthermore, Bidirectional RNN process 
built out versions of each approach above with bidirectional RNNs (BiLSTM, BiGRU, respectively), where
the input sequence
is processed
in both forward and backward directions. This enrichment enables the above technologies to capture the past and future context of the input sequence---yielding a significant advantage over the standard RNN formalism. 
%Thus, the Bidirectional RNNs can capture more information about the input sequence than a standard RNN.

\paragraph{Hyperparameters.}
We used grid-search optimization to derive the optimal parameters 
%of 
for
each method. For consistency, we used the same experimental settings for all models with 10-fold cross-validation, reporting the average score. 
%A v
Varying length
%s of 
posts are padded and trained for 150 epochs with early stopping, and
%with a 
patience of 20 epochs. Thus, we set hyperparameters for our experiments with transformer-based models as $H$ = $256$, $O$ = Adam, learning rate = 1$\times 10^{-3}$, and batch size $128$.%,
% \subsection{Performance Measures}
%\subsection{Evaluation Metrics}

%We used grid-search optimization to derive the optimal parameters of each baseline and used precision, recall, and F1-score as evaluation metrics.
% \subsection{Classifiers}
\subsection{Experimental Results}
Table~\ref{resultss} reports precision, recall and f-score for all classifiers, resulting in non-reliable Accuracy for real-time use. Word2vec achieved the lowest Accuracy of 0.64. We postulate this low performance because word2vec is unable to capture contextual information. GloVe + GRU, a state-of-the-art deep learning model, achieved the highest performance among all recurrent neural network models, counterparts with an F1-score and Accuracy of 0.77 and 0.78, respectively.

\begin{table}[]
\centering
% \vspace{-0.35cm}
\caption{Main results: Comparison of SOTA baselines. Score of each metric is averaged over 10-folds.}
\label{resultss}
\begin{tabular}{p{3cm}|p{1cm}p{1cm}p{1cm}|p{1cm}p{1cm}p{1cm}| p{2cm}}
 \midrule \midrule
Model      & \multicolumn{3}{c}{Absent}  & \multicolumn{3}{c}{Present} & Accuracy \\
% & \multicolumn{6}{6}{\midrule}  &\\
\midrule
& P & R & F & P & R & F & \\ \midrule \midrule
Word2Vec + RF & 0.61 & 0.55 & 0.58  &  0.66 & 0.72 & 0.69 & 0.64 \\% 0.27
GloVe + LSTM &0.60 & 0.82 & 0.70 & 0.81 & 0.58 & 0.68 & 0.69  \\
GloVe+BiLSTM & 0.80 & 0.59 & 0.68 & 0.74 & 0.89 & 0.81 & 0.76 \\

GloVe + GRU &  0.72 & 0.80 & 0.75 & 0.83 & 0.76 & 0.79 &  0.77   \\
% GloVe+ BiGRU & 0.64 & 0.54 & 0.58 & 0.72 &	0.80 &	0.76 & 0.70 \\
% ELMo  & 0.59     & 0.57      & 0.60    \\
Glove + BiGRU &  0.70 & 0.84 & 0.76 & 0.85 & 0.73 & 0.79 &  0.78  \\
% ALBERT & 0.63     & 0.58      & 0.61           \\\midrule \midrule
% DistilBERT &           &        &         \\
% XLNET      &           &        &         \\  
\midrule \midrule
\end{tabular}
\vspace{-0.35cm}
\end{table}

\begin{table}[]
\centering
% \vspace{-0.35cm}
\caption{Performance evaluation of explanations obtained through LIME}
\label{resultss1}
\begin{tabular}{p{2cm}|p{2cm}p{2cm}p{2cm}}
 \midrule \midrule
Model     & ROUGE-1 P & ROUGE-2 R & ROUGE-1 F   \\ \midrule \midrule
LSTM & 0.50 & 0.12 & 0.18\\
BiLSTM & 0.58  & 0.15 & 0.22\\

GRU & 0.53 & 0.14 & 0.21 \\
% GloVe+ BiGRU & 0.64 & 0.54 & 0.58 & 0.72 &	0.80 &	0.76 & 0.70 \\
% ELMo  & 0.59     & 0.57      & 0.60    \\
BiGRU & 0.55 & 0.14 & 0.21  \\
% ALBERT & 0.63     & 0.58      & 0.61           \\\midrule \midrule
% DistilBERT &           &        &         \\
% XLNET      &           &        &         \\  
\midrule \midrule
\end{tabular}
% \vspace{-0.35cm}
\end{table}

We further examine the explanations for Recurrent neural network models through Local Interpretable Model-Agnostic Explanations (LIME). LIME provide a human-understandable explanation of how the model arrived at its prediction of lonesomneness in a given text. We further use ROUGE-1 scores to validate the explanations obtained through LIME, over all positive samples in test data with explainable text-spans in ground-truth of \textsc{LonXplain} (see Table~\ref{resultss1}). We observe that all 
%the 
explanations are comparable and 
%achieves 
achieve
high precision as compared to recall. In the near future, we plan 
%for frame
to formulate 
better explainable approaches by incorporating clinical questionnaires in language models. Consider the following text T1:
\begin{quote}
    T1: What bothers me is the soul crushing \textcolor{red}{loneliness}, i haven't had a \textcolor{red}{girlfriend} in years and I haven't been \textcolor{blue}{physically touched} in what seems like forever. I spend all day in a \textcolor{blue}{shitty} \textcolor{brown}{\textbf{little side room by myself}} writing and hardly see \textcolor{blue}{hide} nor hair of another person besides my dad most of the time. I'm \textcolor{blue}{pretty} done with it all to be \textcolor{blue}{honest}, I don't really see any reason to \textcolor{blue}{continue living like} this.
\end{quote}
BiGRU decides label 1 for T1 with 0.96 prediction probability, highlighting the text-spans: (i) focused by BiGRU for making decision (blue + red colored text), (ii) marked as explanations in the ground truth of \textsc{LonXplain} (red colored text), and (iii) missed text-spans by BiGRU (brown colored text). 

\section{Conclusion and Future Work}

We present \textsc{LonXplain}, a new dataset for identifying lonesomeness through human-annotated extractive explanations from Reddit posts, consisting of 3,522 English Reddit posts labeled across binary labels. In future work, we plan to enhance the dataset with more samples and develop new models tailored explicitly to lonesomeness detection. 

The implications of our work %have 
are the potential to improve public health surveillance and to support other health applications that would benefit from the detection of lonesomeness. 
%automatically identifying lonesomeness 
Automatic detection of lonesomeness
in the posts at early stage of mental health issues
%, thereby
has the potential for
preventing prospective chronic diseases. We %frame 
define
annotation guidelines based on three clinical questionnaires. 
If
%, which if 
accommodated as external knowledge from a lexical resource, the outcome of our study
%shall 
has the potential to improve 
%the 
existing classifiers. We keep this idea as an open research direction.

%\muskan{Add the need of understanding the perspective of a user for KiL through the higher layers of NLP analysis discourse analysis and pragmatics. }

% The explainable AI models developed for lonesomeness classification is applicable to determine social media at-risk users with suicidal prospects. The consequence of loneliness derives the need of early mitigation of deteriorating mental health. 

\section*{Ethical Considerations and Broader Impact}
We emphasize that the sensitive nature of our work necessitates that we use publicly available Reddit posts in a purely observational manner. This research intends to improve public health surveillance and other health applications that automatically identify lonesomeness on Reddit. To adhere to privacy constraints, we do not disclose any personal information such as demographics, location, and personal details of social media user while making \textsc{LonXplain} publicly available~\cite{zirikly2022explaining}. The annotations scheme is carried out under the observation of a senior clinical psychologist, a rehabilitation 
%councillor 
counselor,  %BJD: I really don't think you mean councillor
and a social NLP expert. This research is purely observational and we do not claim any solution for clinical diagnosis at this stage~\cite{gaur2022iseeq}. Reddit posts might subject to biased demographics such as race, location and gender of a user. Therefore, we do not claim \textit{diversity} in our dataset. Our dataset is susceptible to the prejudices and biases of our student annotators. There will be no ethical issues or legal impact with our dataset and is subject to IRB approval.

\section*{Acknowledgement}
We would like to sincerely thank the postgraduate student annotators, Ritika Bhardwaj, Astha Jain, and Amrit Chadha, for their dedicated work in the annotation process. We are grateful to Veena Krishnan, a senior clinical psychologist, and Ruchi Joshi, a rehabilitation counselor, for their unwavering support during the project. Furthermore, we would like to express our heartfelt appreciation to Prof. Sunghwan Sohn for consistently guiding and supporting us.

\bibliographystyle{splncs04}
\bibliography{mybibliography}

\begin{thebibliography}{10}
\providecommand{\url}[1]{\texttt{#1}}
\providecommand{\urlprefix}{URL }
\providecommand{\doi}[1]{https://doi.org/#1}

\bibitem{gaur2022iseeq}
Gaur~et al., M.: Iseeq: Information seeking question generation using dynamic
  meta-information retrieval and knowledge graphs (2022)

\bibitem{ghosh2022no}
Ghosh~et al., S.: Am i no good? towards detecting perceived burdensomeness and
  thwarted belongingness from suicide notes. IJCAI  (2022)

\bibitem{asher1984children}
Asher, S., Wheeler, V.: Children’s loneliness and social dissatisfaction
  scale. Child Development  \textbf{55}(4),  1456--1464 (1984)

\bibitem{badal2021words}
Badal, V.D., Nebeker, C., Shinkawa, K., Yamada, Y., Rentscher, K.E., Kim, H.C.,
  Lee, E.E.: Do words matter? detecting social isolation and loneliness in
  older adults using natural language processing. Frontiers in psychiatry
  \textbf{12} (2021)

\bibitem{baumeister2017need}
Baumeister, R.F., Leary, M.R.: The need to belong: Desire for interpersonal
  attachments as a fundamental human motivation. Interpersonal development pp.
  57--89 (2017)

\bibitem{becker2022surviving}
Becker, W.J., Belkin, L.Y., Tuskey, S.E., Conroy, S.A.: Surviving remotely: How
  job control and loneliness during a forced shift to remote work impacted
  employee work behaviors and well-being. Human Resource Management  (2022)

\bibitem{cacioppo2006loneliness}
Cacioppo, J.T., Hughes, M.E., Waite, L.J., Hawkley, L.C., Thisted, R.A.:
  Loneliness as a specific risk factor for depressive symptoms: cross-sectional
  and longitudinal analyses. Psychology and aging  \textbf{21}(1), ~140 (2006)

\bibitem{donovan2020social}
Donovan, N.J., Blazer, D.: Social isolation and loneliness in older adults:
  review and commentary of a national academies report. The American Journal of
  Geriatric Psychiatry  \textbf{28}(12),  1233--1244 (2020)

\bibitem{garg2023mental}
Garg, M.: Mental health analysis in social media posts: A survey. Archives of
  Computational Methods in Engineering pp. 1--24 (2023)

\bibitem{garg2022cams}
Garg, M., Saxena, C., Krishnan, V., Joshi, R., Saha, S., Mago, V., Dorr, B.J.:
  Cams: An annotated corpus for causal analysis of mental health issues in
  social media posts. In: Language Resources Evaluation Conference (LREC)
  (2022)

\bibitem{ghosh2022persona}
Ghosh, S., Maurya, D.K., Ekbal, A., Bhattacharyya, P.: Em-persona:
  Emotion-assisted deep neural framework for personality subtyping from suicide
  notes. In: Proceedings of the 29th International Conference on Computational
  Linguistics. pp. 1098--1105 (2022)

\bibitem{grootendorst2020keybert}
Grootendorst, M.: Keybert: Minimal keyword extraction with bert. Zenodo  (2020)

\bibitem{kivran2014understanding}
Kivran-Swaine, F., Ting, J., Brubaker, J., Teodoro, R., Naaman, M.:
  Understanding loneliness in social awareness streams: Expressions and
  responses. In: Proceedings of the International AAAI Conference on Web and
  Social Media. vol.~8, pp. 256--265 (2014)

\bibitem{mahoney2019feeling}
Mahoney, J., Le~Moignan, E., Long, K., Wilson, M., Barnett, J., Vines, J.,
  Lawson, S.: Feeling alone among 317 million others: Disclosures of loneliness
  on twitter. Computers in Human Behavior  \textbf{98},  20--30 (2019)

\bibitem{NIPS2013_5021}
Mikolov, T., Sutskever, I., Chen, K., Corrado, G.S., Dean, J.: Distributed
  representations of words and phrases and their compositionality. In: Burges,
  C.J.C., Bottou, L., Welling, M., Ghahramani, Z., Weinberger, K.Q. (eds.)
  Advances in Neural Information Processing Systems 26, pp. 3111--3119. Curran
  Associates, Inc. (2013),
  \url{http://papers.nips.cc/paper/5021-distributed-representations-of-words-and-phrases-and-their-compositionality.pdf}

\bibitem{nemecek2020loneliness}
Nemecek, D.: Loneliness and the workplace: 2020 us report. Cigna, January
  (2020)

\bibitem{nguyen2018comparing}
Nguyen, D.: Comparing automatic and human evaluation of local explanations for
  text classification. In: Proceedings of the 2018 Conference of the North
  American Chapter of the Association for Computational Linguistics: Human
  Language Technologies, Volume 1 (Long Papers). pp. 1069--1078 (2018)

\bibitem{nottage2022loneliness}
Nottage, M.K., Oei, N.Y., Wolters, N., Klein, A., Van~der Heijde, C.M., Vonk,
  P., Wiers, R.W., Koelen, J.: Loneliness mediates the association between
  insecure attachment and mental health among university students. Personality
  and Individual Differences  \textbf{185},  111233 (2022)

\bibitem{penning2014measuring}
Penning, M.J., Liu, G., Chou, P.H.B.: Measuring loneliness among middle-aged
  and older adults: The ucla and de jong gierveld loneliness scales. Social
  Indicators Research  \textbf{118},  1147--1166 (2014)

\bibitem{pennington2014glove}
Pennington, J., Socher, R., Manning, C.: Glove: Global vectors for word
  representation. In: Proceedings of the 2014 conference on empirical methods
  in natural language processing (EMNLP). pp. 1532--1543 (2014)

\bibitem{russell1980revised}
Russell, D., Peplau, L.A., Cutrona, C.E.: The revised ucla loneliness scale:
  concurrent and discriminant validity evidence. Journal of personality and
  social psychology  \textbf{39}(3), ~472 (1980)

\bibitem{weiss1975loneliness}
Weiss, R.: Loneliness: The experience of emotional and social isolation. MIT
  press (1975)

\bibitem{wilkinson2016fair}
Wilkinson, M.D., Dumontier, M., Aalbersberg, I.J., Appleton, G., Axton, M.,
  Baak, A., Blomberg, N., Boiten, J.W., da~Silva~Santos, L.B., Bourne, P.E.,
  et~al.: The fair guiding principles for scientific data management and
  stewardship. Scientific data  \textbf{3}(1), ~1--9 (2016)

\bibitem{zirikly2022explaining}
Zirikly, A., Dredze, M.: Explaining models of mental health via clinically
  grounded auxiliary tasks. In: CLPsych (2022)

\end{thebibliography}

\end{document}